\documentclass[10pt, a4paper]{article}

\usepackage{lrec-coling2024} 
\usepackage{helvet}
   
\usepackage{natbib}
\makeatletter
\def\@mb@citenamelist{cite,citep,citet,citealp,citealt,citepalias,citetalias}
\makeatother

\usepackage{graphicx}
\usepackage{tabularx}
\usepackage{soul}
\usepackage{spverbatim}

\usepackage{xcolor}
\usepackage{hyperref}
 \definecolor{darkblue}{rgb}{0, 0, 0.5}
  \hypersetup{colorlinks=true, citecolor=darkblue, linkcolor=darkblue, urlcolor=darkblue}

\usepackage{color}

\usepackage{array}
\newcolumntype{L}[1]{>{\raggedright\let\newline\\\arraybackslash\hspace{0pt}}m{#1}}

\title{Enhancing Emotion Prediction in News Headlines: Insights from ChatGPT and Seq2Seq Models for Free-Text Generation}

\name{Ge Gao*, Jongin Kim*, Sejin Paik*, Ekaterina Novozhilova*, Yi Liu*, Sarah Bonna*, \\ {\bf \large Margrit Betke*, Derry Wijaya*$^{\dagger}$}}

\address{*Boston University, $^{\dagger}$Monash University Indonesia\\
          \{ggao02, jongin, sejin, ekaterin, yliu2702, sbonna, betke, wijaya\}@bu.edu\\}

\abstract{
Predicting emotions elicited by news headlines can be challenging as the task is largely influenced by the varying nature of people's interpretations and backgrounds. Previous works have explored classifying discrete emotions directly from news headlines. 
We provide a different approach to tackling this problem by utilizing people's explanations of their emotion, written in free-text, on how they feel after reading a news headline. Using the dataset BU-NEmo$^+$ \cite{gao-etal-2022-prediction}, we found that for emotion classification, the free-text  explanations have a strong correlation with the dominant emotion elicited by the headlines. The free-text explanations also contain more sentimental context than the news headlines alone and can serve as a better input to emotion classification models. 
Therefore, in this work we explored generating emotion explanations from headlines by training a sequence-to-sequence transformer model and by using pretrained large language model, ChatGPT (GPT-4). We then used the generated emotion explanations for emotion classification. In addition, we also experimented with training the pretrained T5 model for the intermediate task of explanation generation before fine-tuning it for emotion classification. Using McNemar's significance test, methods that incorporate GPT-generated free-text emotion explanations demonstrated significant improvement (P-value < 0.05) in emotion classification from headlines, compared to methods that only use headlines. This underscores the value of using intermediate free-text explanations for emotion prediction tasks with headlines.  
\\ \newline  }
\begin{document}
\maketitleabstract
\section{Introduction}
Emotion classification from text is an important task in natural language processing. Extracting this information has multiple applications \cite{alhuzali} 
including health \cite{health1, health2, health3}, customer behavior studies \cite{customerbehavior}, understanding public opinion regarding social policies and political decisions \cite{kanojia2023applications}, and profiling based on user characteristics \cite{userprofile}. Moreover, previous study \cite{emotion-from-text} has shown that narrative texts are prone to contain emotional content and in the case of text-to-speech synthesis, speakers can more effectively deliver speech and make it particularly appealing when they have knowledge of its emotional content 
that enables them to modify the pitch, intensity, and duration cues in their speech.

In the domain of news media, the task of classifying the emotion of news content has been receiving increasing attention. 
Understanding emotions can aid news consumers in grouping news articles into emotion categories \cite{jia2009} and help with text-to-speech synthesis for news producers. In addition, identifying emotions elicited by different news content can help news reporters deliver content that is more engaging to the audience. 

Most research on understanding emotions from news text have focused on the task of sentiment analysis and extracting the conveyed emotions in  text from the writer’s intent \cite{vasava, 9066983}. However, it is informative to be able to predict readers’ emotional reactions after exposure to news articles \cite{lin2007}. \citet{4740453} attempts to classify emotions of online news articles from the reader's perspective, and has pointed out several applications of this. One application is a "reader-emotion classification" that can be integrated into a search engine on the web. Such application will allow users to retrieve documents that contain the relevant information they are searching for while "instilling proper emotions", depending on their search keywords.

In this work, we are interested in predicting emotions from news headlines \cite{desh2012, jia2009, gao-etal-2022-prediction, Patil2013UseOP, stock-news-emotion} since headlines are meant to draw readers’ attention and provoke emotions \cite{desh2012}. 
Most previous studies on this have approached emotion classification relying only on the headlines themselves. However, the emotional information contained in the headlines is sparse and not always subjective as some headlines are designed to be more sensational \cite{sensational}. \citet{freetext-misinfo} highlighted that it is hard to capture the nuanced emotional reactions towards news headlines with just the headline text and turn to the free-text explanations of readers reactions. Their study shows that neural networks can predict readers' reactions towards news headline, distinguishing real news from misinformation. These findings demonstrate that the free-text emotion explanations contain more context about the readers' perceptions and emotional reactions towards the news. Therefore, we hypothesize that using free-text explanations of emotional reactions towards the headlines in emotion classification is better than only relying on the headlines.

\subsection{Text Generation for Emotion Classification}
We want to explore if we can harness the rich emotional context of the free-text explanation of the emotional reactions when in practice, at inference time we will only have access to the news headlines. Therefore, we turn to generating these free-text emotion explanations from the headlines. Text generation is a natural language processing task that received increasing attention with the appearance of Transformer architecture models including GPT variations \cite{gpt} and T5 \cite{raffel2020exploring}. In recent years, text generation has shown promising results in numerous fields of application \cite{text-generation-survey} including machine translation \cite{transformer, machine-translation}, text summarization \cite{text-summarization}, and medical applications \cite{medical-augment-text-generation}. More recently, the GPT-based generative chatbot, ChatGPT, gained spotlight as it demonstrated high quality performance at question answering and text generation, including news article summarization \cite{wang2023chatgpt}. In our study, we aim to explore the use of these large pretrained language models to generate free-text emotion explanations based on headlines and see whether they can help with emotion classification.

\section{Method}
\subsection{Data}

\begin{figure}[t]
\centering
\includegraphics[width=1\columnwidth]{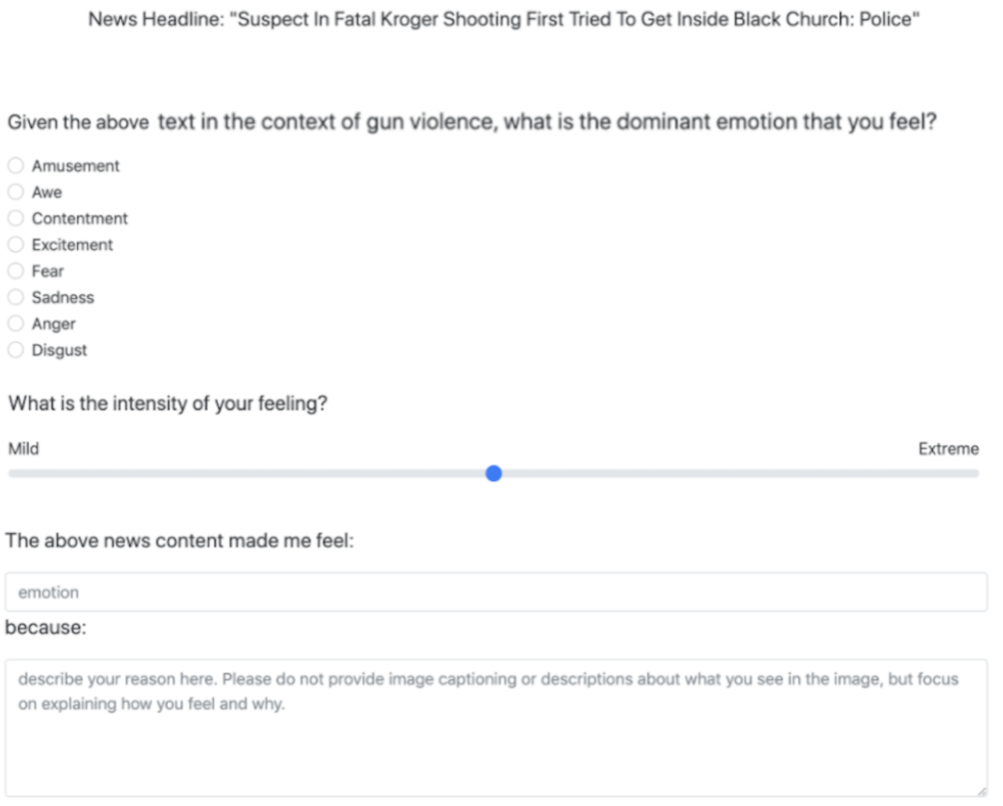}
\caption{The interface for the gun violence emotion annotation crowdsourcing experiment for a single news headline.}
\label{sona_interface}
\end{figure}

For our study of predicting the emotion from news consumers after reading news headlines, we used the BU-NEmo$^{+}$ \cite{gao-etal-2022-prediction} dataset which contains 1297 headline and lead news image pairings for gun violence news. Starting with the Gun Violence Frame Corpus (GVFC) dataset which applies framing predictions on news headlines and their lead images related to gun violence \cite{liu-etal-2019-detecting, akyurek-etal-2020-multi, tourni-etal-2021-detecting-frames}, BU-NEmo \cite{reardon} extended upon it by collecting emotion annotations from a crowdsourcing experiment on Amazon Mechanical Turk (MTurk) for 320 gun violence related news headlines. There are 10 emotion annotations for each headline and each annotation consists of 1) an emotion label from the 8 categories (Amusement, Awe, Contentment, Excitement, Fear, Sadness, Anger, and Disgust) \cite{Mikels}; 2) the intensity of the emotion on a scale of 1-5; and 3) a short free text explaining why the annotator feels the selected emotion, as shown in Figure \ref{sona_interface}. The experiment was repeated for 3 modalities: Text only (T), Image only (I), and Text+Image (TI), respectively corresponding to when the annotators were presented with only the headline, only the image, and the headline together with the image. BU-NEmo$^+$ expanded the original BU-NEmo dataset to include emotion annotations for 1297 headlines. We only used the Text only modality of the data for our study.

For our analysis, we used the full BU-NEmo$^+$ and its subset, BU-NEmo$^+$-CR, containing 365 news items that show a clear emotional agreement. \citet{gao-etal-2022-prediction} highlighted that people have varying emotion reactions towards the same news content in the BU-NEmo$^{+}$ dataset, and it is a challenging task to select a single ground truth emotion label for a given news item. This adds a significant layer of complexity and variance to the emotion prediction task with single label classification.

Moreover, due to the nature of the gun violence news \cite{doi:10.1080/15205436.2021.1898644,doi:10.1177/00936502231151555}, the dataset is imbalanced towards the negative emotions. Therefore, it is important to take the majority baseline into consideration during evaluations of the models' performances.

\subsection{Emotion Prediction from Explanations} \label{section-cee}
\begin{figure}[t]
\centering
\includegraphics[width=1\columnwidth]{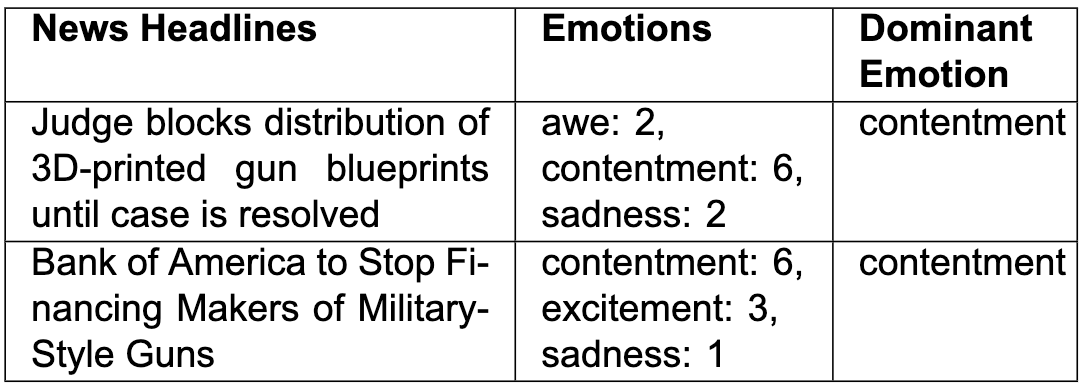}
\caption{Examples of headlines that do not contain the sentiment information of the reactions.}
\label{tab:no_sentiment}
\end{figure}

The BU-NEmo$^+$ study used headlines as the input to the emotion classification model. However, relying solely on the headline comes with the disadvantage that headlines often carry limited information about the news. Furthermore, predicting the emotional reaction from headlines is especially difficult since the headline text does not usually contain the sentiment information of the emotional reaction, as shown by examples in Figure \ref{tab:no_sentiment}. To tackle this problem, we turn to the emotion explanations. For the first example headline that is about blocking 3D-printed gun blueprints (Figure \ref{tab:no_sentiment}), the emotion explanations justifying why annotators feel a certain emotion towards the news are: 
\begin{quote}

\begin{tiny}
\begin{spverbatim}
1. for now i am content but intrigued to see the next steps
2. the distribution of 3D-printed gun blueprints should not happen
3. I think the judge did a right thing; that is good of the judge
4. Seems like a strange call, but not one that I have enough knowledge to properly have an opinion on
5. I'm a fan of strict gun control
6. It seems like a fair solution for the time
7. I can't believe that there is even a discussion about distributing blueprints to 3D-print guns
8. It is already far too easy to gain access to guns in this country
9. I am confused as to why this would happen
10. because why are there printed gun blue prints in the first place
\end{spverbatim}    
\end{tiny}

\end{quote}

It is easy to observe that the emotion explanation has the advantage of containing the desired emotional information that is easier to classify than the headline. 

We compiled the emotion explanations from BU-NEmo$^+$ and refer to this dataset of free-text emotion explanations and their emotion labels as \textit{EE} (Emotion Explanations). We also created another dataset by concatenating all 10 emotion explanations in BU-NEmo$^+$-CR for each news which we refer to as the \textit{CEE} (Concatenated Emotion Explanations). With the constructed corpus and the emotion labels, we trained a RoBERTa \cite{roberta} classification model
\footnote{https://huggingface.co/roberta-base}. 
We refer to this model as \textbf{CEE}. We followed the setup from BU-NEmo$^+$'s experiments \cite{gao-etal-2022-prediction} and split the data into training / validation / test sets in a ratio of 0.5:0.25:0.25. For a fair comparison, we also included the single label emotion classification with only the headline text, 
a model which we refer to as \textbf{Headline}, and compared it against our results.

Furthermore, due to the diversifying nature of the emotional reaction, in addition to the exact match accuracy for the dominant emotion for a given news headline, we included the top-2 emotion accuracy. We calculated the top-2 emotion accuracy by checking whether the ground truth dominant emotion is the same as either one of the top-2 predicted emotions.  

In our experiment, we found that over 74\% of news headlines in the dataset exhibit a second most dominant emotion with a frequency of at least 2 (out of 10 annotations for each headline). As a result, we believe that measuring the accuracy of the top-2 emotions is a more representative metric than the exact match accuracy, capturing the diverse nature of emotional responses.

However, we would like to emphasize that this task of predicting emotions from explanations is easier since at test time, the \textbf{CEE} model is evaluated on concatenated human-generated free-text emotion explanations. For the rest of the experiments in our study, at inference time we will only have access to the headlines. \textbf{CEE} is our proof-of-concept of whether free-text emotion explanations are helpful in emotion classification. Therefore, the performance of \textbf{CEE} serves as an upper bound in our study.

As shown in the results section \ref{freetext_emotion_section}, using free-text emotion explanations produces a significant improvement in emotion classification than just relying on headlines. This finding led us to explore whether we can further harness the emotion explanations to predict people's emotional reactions towards news content.

\subsection{Seq2Seq: Headlines2Explanations}
\begin{figure}[t]
\centering
\includegraphics[width=1\columnwidth]{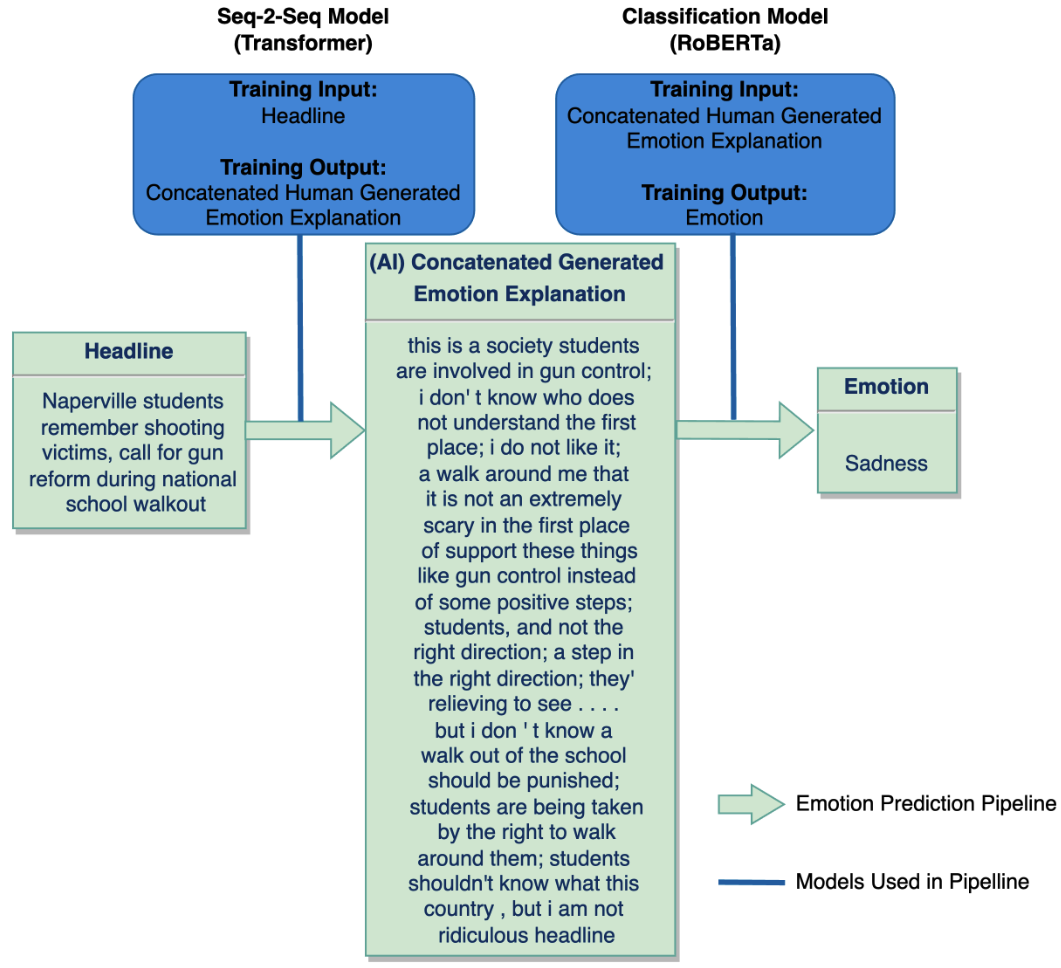} 
\caption{\textbf{CEE-T}. Sequence-to-sequence emotion classification architecture. }
\label{transformer}
\end{figure}

In practice, at inference time, we wanted to utilize the rich emotional context in the emotion explanations but we will only have access to headlines. Therefore, we constructed a seq2seq (sequence-to-sequence) architecture to map a headline to a collection of emotion explanations. The ground truth emotion explanations that the headlines map to were the \textit{CEE} corpus we constructed. 
We trained a sequence-to-sequence transformer \cite{transformer} model and fed the generated emotion explanations into the RoBERTa classification model trained on the BU-NEmo$^+$-CR for emotion classification (shown in Figure \ref{transformer}). We call this model \textbf{CEE-T} (Transformer-generated Concatenated Emotion Explanations). We use the task of generating free-text emotion explanation to draw a parallel to the intermediate reasoning steps that humans perform when conducting a task.

\subsection{Intermediate Task Transfer Learning for Emotion Prediction}
We constructed a model utilizing a seq2seq architecture to generate a collection of emotional explanations that corresponded to headlines in order to leverage the emotional cues presented in the emotional explanations. However, due to the limited size of our training dataset (\textit{CEE}) for this task, there were limitations in generating free-text emotional explanations that mimic human writing 
in terms of sentence completeness (as shown in the example in Figure \ref{transformer}).

\begin{figure}[t]
\centering
\includegraphics[width=1\columnwidth]{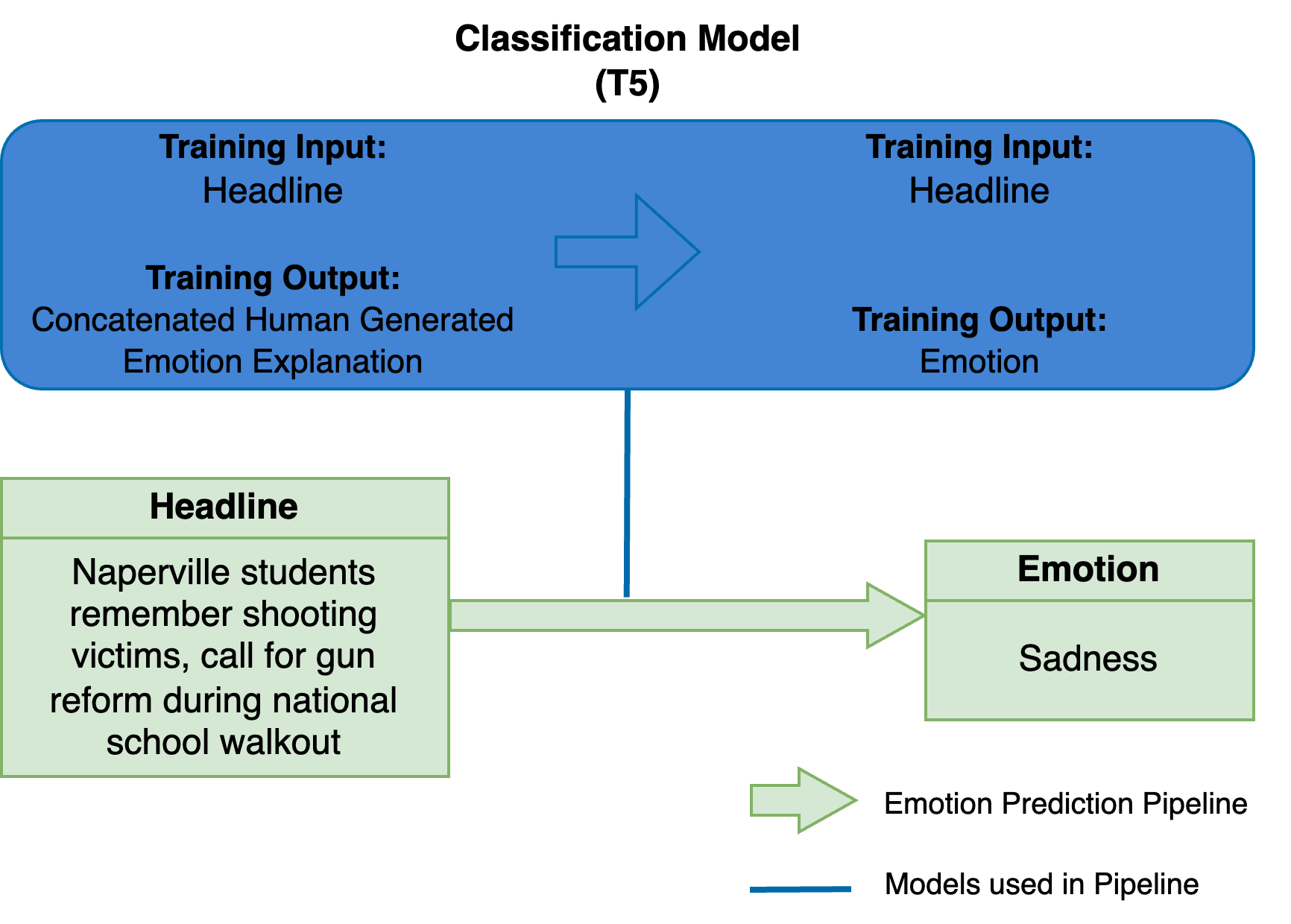} 
\caption{\textbf{T5 with transfer learning}. Pretrained T5 is first fine-tuned on headline to concatenated emotion explanation pairs and subsequently on headline to emotion pairs.}
\label{t5-repurpose}
\end{figure}

Therefore, instead of directly generating free-text emotional explanations and using the generated explanations to predict emotions, we treated the free-text emotion explanation generation task as an intermediate or related task for emotion prediction and performed intermediate-task transfer learning \cite{phang2019sentence}. Implementation wise, this means fine-tuning a pre-trained T5 \cite{raffel2020exploring} model\footnote{https://huggingface.co/google-t5/t5-base} on our headline-to-explanation corpus (\textit{CEE}), and subsequently fine-tuning the same model for the headline-to-emotion prediction task as shown in Figure \ref{t5-repurpose}. The proposed approach does not rely on emotion explanations as direct inputs for predicting emotions. However, through intermediate-task training process, the model parameters can be updated to learn and capture the relationship between the headlines and the emotion explanations written by humans. This learned representation can then be transferred to the target task of emotion prediction, making the training process more effective.

\subsection{Explanation Generation with ChatGPT for Emotion Prediction} \label{baseline-1}

In this section, we describe a way in which emotion explanations generated by the Large Language Model ChatGPT (GPT-4 based) for a news headline can be utilized for emotion prediction. We experimented with different prompting methods (zero-shot and few-shot) for emotion explanation generations with ChatGPT. We also evaluated ChatGPT for a dominant emotion prediction task (\textbf{Baseline 1}) by asking ChatGPT to directly generate the top-2 dominant emotions that readers might feel after reading the headlines using the following prompt: 

\begin{tiny}
\begin{spverbatim}
For each of the headlines below, what are the top 2 dominant emotions (must choose from these 8 emotions: Amusement, Awe, Contentment, Excitement, Fear, Sadness, Anger, and Disgust) that the reader might feel after reading the headlines? The dominant emotion is the emotion most commonly experienced among/by the majority of population/individuals who read the headline. Please only return a list of 91 dominant emotions tuples that the readers might feel corresponding to each of the headlines. For each tuple in the list, the first element should be the most likely dominant emotion and the second element should be the second most likely dominant emotion. Here are the headlines: [...]
\end{spverbatim}
\end{tiny}

\subsubsection{Zero-Shot Explanation Generations} \label{zero-shot-generations}

\begin{figure}[t]
\centering
\includegraphics[width=1\columnwidth]{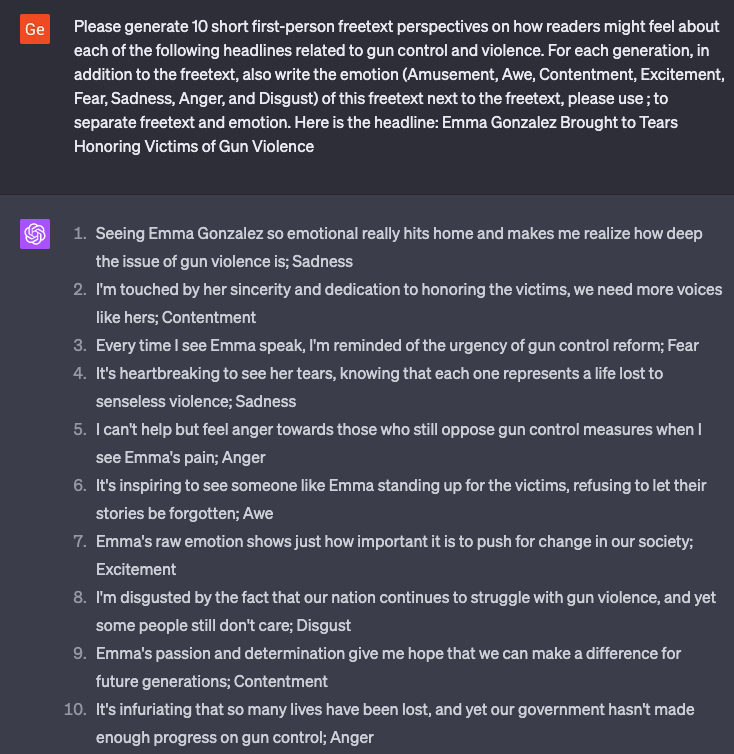} 
\caption{Text prompt and generation format of the zero-shot free-text emotion explanation generation with ChatGPT. In the generation, each free-text explanation is followed with an emotion label after the semicolon. }
\label{chatgpt-prompt-zs}
\end{figure}

In order to leverage ChatGPT-generated emotion explanations, we asked ChatGPT to generate 10 short first-person free-text perspectives given a headline in the test set of \textit{CEE}. Moreover, we also asked ChatGPT to generate the emotion label for each of its free-text emotion explanation generation, as shown in Figure \ref{chatgpt-prompt-zs}. These ChatGPT-generated emotion labels are used as an additional emotion classification baseline (\textbf{Baseline 2}) for evaluation. 

\subsubsection{Few-Shot Explanation Generations}
To mitigate biases and narrow perspectives \cite{gpt-bias} of the large language model, we provided ChatGPT with example headlines and their corresponding emotion explanations from the train set of \textit{CEE}.
For our few-shot approach, the first part of the text prompt was the same as the zero-shot prompt. For the second part of the text prompt, we provided ChatGPT with examples (headline with human annotated emotions and free-text explanations) to learn from. 

In addition, the headlines in the BU-NEmo$^+$-CR dataset originally came from the GVFC \cite{liu-etal-2019-detecting} dataset which contained framing information. Each headline has an associated frame label from 9 framing categories: 2nd Amendment, Gun Control/Regulation, Politics, Mental Health, School/Public Space Safety, Race/Ethnicity, Public Opinion, Society/Culture, and Economic Consequences. To evaluate the impact of these frames on emotion generations, we randomly selected from the train set of \textit{CEE}, a headline from each of these 9 framing categories, and gave the selected headlines and their associated emotion and emotion explanations as few-shot examples to ChatGPT. Below is the format of our few shot text prompt:

\begin{tiny}
\begin{spverbatim}

Please generate 10 short first-person freetext perspectives on how readers might feel about each of the following headlines related to gun control and violence. For each generation, in addition to the freetext, also write the emotion (Amusement, Awe, Contentment, Excitement, Fear, Sadness, Anger, and Disgust) of this freetext next to the freetext, please use ; to separate freetext and emotion. 
Here are 9 examples that I want you to learn from, mimic the content and learn from people's freetext explanations examples: 

headline 1: Marchers Should Demand Second Amendment Repeal, Says Former Supreme Court Justice Stevens

the constitution is the backbone of the country, without it, we are not America; anger
I'm not interested in this kind of news.; fear
Motivated to never join a march like this; contentment
as a former Supreme Court Justice, he would know if there was enough evidence that could determine the amendment unconstitutional.; contentment
Repeal takes time and cause more problem. I believe there is a wiser way to control gun violence step by step. ; fear
Is this saying that it's bad because they think they are very influential or bad because they are not influential?; contentment
i agree with former justice stevens' statement but it is sad that most don't and allow gun violence to run rampant in our society; sadness
The issue is one of the most complex in society; sadness
i dont think that repealing the second amendment is going to fix any problems.; contentment
......

headline 9: The NRA Says It's Suffered 'Tens Of Millions Of Dollars' Of Harm Since Parkland
...

Here is the headline for you to generate freetext: Emma Gonzalez Brought to Tears Honoring Victims of Gun Violence

\end{spverbatim}
\end{tiny}

The few-shot ChatGPT generations are in the same format as the zero-shot generations (Figure \ref{chatgpt-prompt-zs}) and also produce the emotion labels next to the emotion explanations for us to evaluate ChatGPT's few-shot emotion classification performance (i.e., few-shot \textbf{Baseline 2}). 
In addition to selecting few-shot examples from each frame, we also select few-shot examples randomly without knowledge of frames and evaluate the performance. 
We also include these two few-shot sampling approaches for \textbf{Baseline 1} (section \ref{baseline-1}), where we ask ChatGPT to directly predict the top-2 dominant emotions given a headline. 


\subsubsection{Emotion Classification Pipelines}

For each type of generation (zero- and few-shot), we explored two pipelines. Both have headlines as inputs and dominant emotions as outputs:

\begin{figure}[t]
\centering
\includegraphics[width=1\columnwidth]{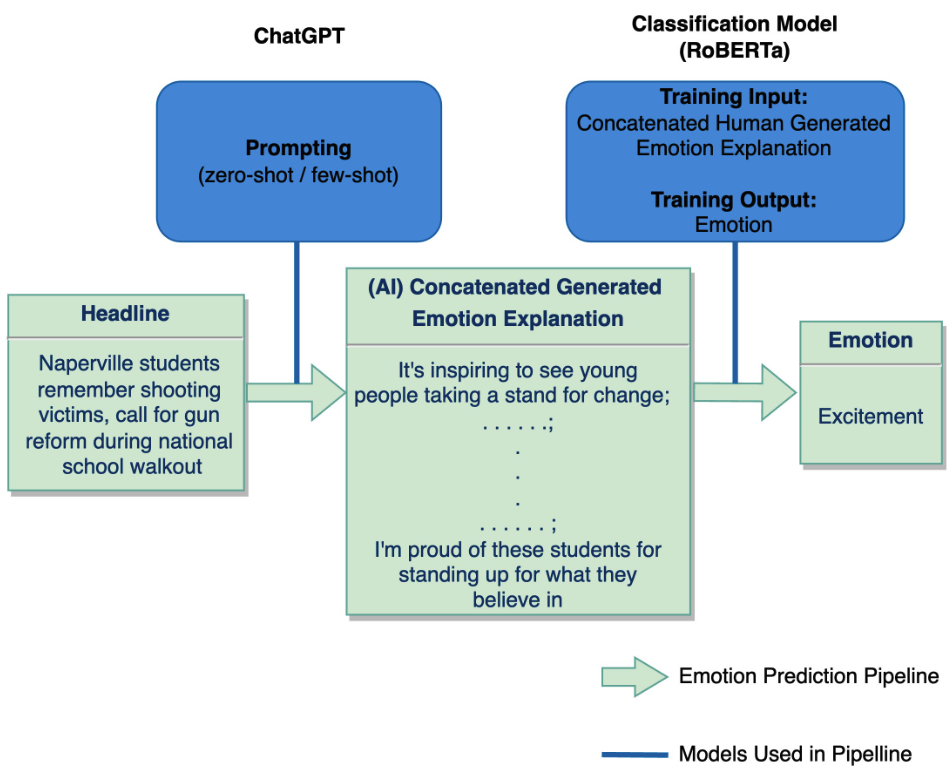} 
\caption{\textbf{CEE-Chat} (zero- and few-shot). Headline to Concatenated Emotion Explanation generation with ChatGPT and emotion classification.}
\label{chatgpt_concat}
\end{figure}

\begin{figure}[t]
\centering
\includegraphics[width=1\columnwidth]{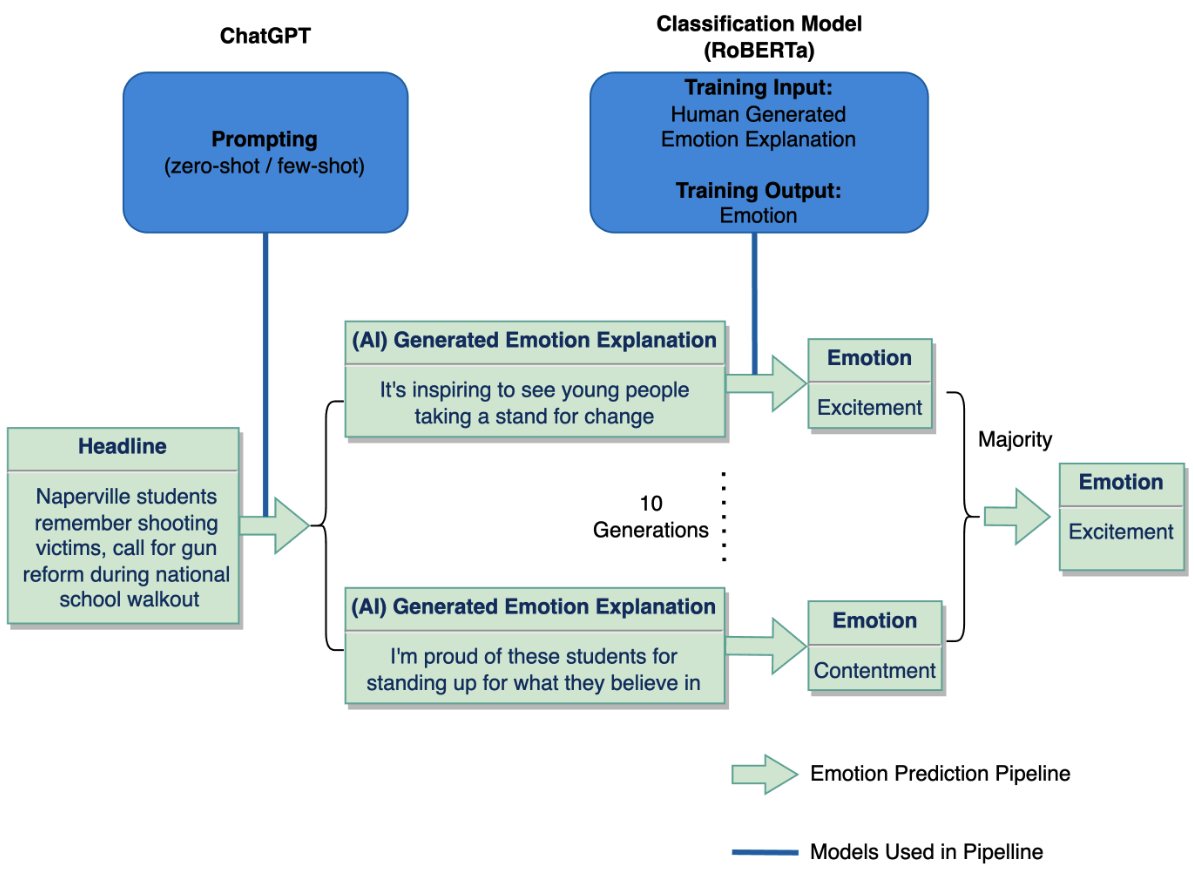}

\caption{\textbf{EE-Chat} (zero-shot and few-shot). Headline to Emotion Explanation generation with ChatGPT and emotion classification.}
\label{chatgpt_singles}
\end{figure}

\begin{enumerate}
  \item As shown in Figure \ref{chatgpt_concat}, we trained the RoBERTa classification model where the input was the Concatenated Emotion Explanations in \textit{CEE} and the output was the dominant emotion. At test time, we used the concatenated ChatGPT-generated emotion explanations to predict the dominant emotion labels. We refer to this model pipeline as \textbf{CEE-Chat}.
  \item Instead of relying on the entire Concatenated Emotion Explanations which may contain multiple sentiments and perspectives for emotion classification, we experimented with emotion classification with individual Emotion Explanation in \textit{EE} and took a majority vote on the predicted emotions to determine the dominant emotion for a given headline, as shown in Figure \ref{chatgpt_singles}. We refer to this model pipeline as \textbf{EE-Chat}.
\end{enumerate}


\section{Results}
\subsection{Emotion Explanations for Emotion Prediction} \label{freetext_emotion_section}
For the single label emotion classification, the majority baseline which represents the percentage of the dominant class for this dataset is 0.42.
As shown from the test time performances in Table \ref{tab:single_labels}, the models using emotion explanations show significant performance improvement in both the exact match and top-2 emotions accuracies for the task of single label emotion classification. The \textbf{CEE} outperformed the \textbf{Headline} model by 10 percent points for the exact match accuracy and 13 percent points for the top-2 accuracy.

\begin{table}[htbp]
    \begin{center}

    \begin{tabular}{ |p{0.18\columnwidth} | p{0.23\columnwidth}| p{0.37\columnwidth}| }
     \hline
     & Top-2 Acc & Exact Match Acc  \\
     \hline
     \textbf{CEE} & 0.84 & 0.68  \\
     \hline
     \textbf{Headline} & 0.71 & 0.58  \\
     \hline
    \end{tabular}
    
    \caption{The exact match and top-2 test accuracies for the single label emotion classifications. The model \textbf{Headline} was trained and evaluated on the headlines and the model \textbf{CEE} on the Concatenated human-generated Emotion Explanations in the \textit{CEE} dataset. Performances are reported as averages across 30 runs. }
    \label{tab:single_labels}
    \end{center}
\end{table}

As mentioned in section \ref{section-cee}, the task of the \textbf{CEE} model here is intrinsically easier and its performance serves as an upper bound when we have human-generated emotion explanations for each headline at inference time. We are more interested in the real-world scenario where at inference time we only have the headlines to predict emotions. Indeed, the main focus of our study is our novel approach in using \textit{generated} free-text emotion explanations to enhance the task of emotion prediction from headlines.

\subsection{Intermediate Task Transfer Learning}

As shown in Table \ref{tab:intermediate}, intermediate-task training on headline-to-explanation generation task boosts the performance of the emotion classification task by 6 percent points. While this T5 model with transfer learning does not surpass the exact match accuracy of 0.58 achieved by \textbf{Headline} model (Table \ref{tab:single_labels}), the results demonstrate that leveraging emotion explanations could enhance the emotion classification performance compared to using the headlines alone. Moreover, the findings show positive transferability between emotion explanation generation and emotion classification tasks.


\begin{table}[htbp]
\centering
\resizebox{\columnwidth}{!}{\begin{tabular}{|c|c|c|}
\hline
 & Top-2 Acc & Exact Match Acc \\
\hline
\textbf{T5 w/o transfer learning} & 0.68 & 0.52 \\
\hline
\textbf{T5 w/ transfer learning} & 0.69 & 0.58 \\
\hline
\end{tabular}
}
\caption{The exact match and top-2 test accuracies for the single label emotion classifications with and without transfer learning. Performances are reported as  averages across 30 runs.}
\label{tab:intermediate}
\end{table}

\subsection{Free Text Explanation Generation for Emotion Classification}

\begin{table}[htbp]
    \begin{center}

    \begin{tabular}{ |p{0.23\columnwidth} | p{0.11\columnwidth}| p{0.22\columnwidth}| p{0.2\columnwidth}|  }
     \hline
     \multicolumn{4}{|c|}{Top-2 Acc} \\
     \hline
     & Zero-shot & Few-shot w/o frames & Few-shot w/ frames \\
     \hline
     \textbf{Baseline 1} & 0.64 & 0.75 & 0.75 \\
     \hline
     \textbf{Baseline 2} & 0.59 & 0.85 & 0.81 \\
     \hline
     \textbf{EE-Chat} & 0.73 & 0.85 & 0.78 \\
     \hline
     \textbf{CEE-Chat} & 0.77 & 0.81 & 0.78 \\
     \hline
     \textbf{CEE-T} & \multicolumn{3}{|c|}{0.61} \\
     \hline
     
    \end{tabular}

    \begin{tabular}{ |p{0.23\columnwidth} | p{0.11\columnwidth}| p{0.22\columnwidth}| p{0.2\columnwidth}|  }
     \hline
     \multicolumn{4}{|c|}{Exact Match Acc} \\
     \hline
     & Zero-shot & Few-shot w/o frames & Few-shot w/ frames \\
     \hline
     \textbf{Baseline 1} & 0.41 & 0.63 & 0.58 \\
     \hline
     \textbf{Baseline 2} & 0.46 & 0.64 & 0.59 \\
     \hline
     \textbf{EE-Chat} & 0.48 & 0.61 & 0.60 \\
     \hline
     \textbf{CEE-Chat} & 0.61 & 0.66 & 0.65 \\
     \hline
     \textbf{CEE-T} & \multicolumn{3}{|c|}{0.47} \\
     \hline
     
    \end{tabular}
    
    \caption{The exact match and top-2 test accuracies for the single label emotion classifications using generated free-text emotion explanations. \textbf{CEE-T} denotes the pipeline in Figure \ref{transformer} while \textbf{EE-Chat} and \textbf{CEE-Chat} stand for the pipelines illustrated in Figure \ref{chatgpt_concat} and Figure \ref{chatgpt_singles} respectively. Performances are averages across 30 runs. }
    \label{tab:chatgpt_zsfs}
    \end{center}
\end{table}

As shown in Table \ref{tab:chatgpt_zsfs}, the emotion classification pipelines using ChatGPT to generate explanations (\textbf{EE-Chat}, \textbf{CEE-Chat}) significantly outperform training our own seq2seq explanation-generation transformer (\textbf{CEE-T}).

When using emotion explanations generated by ChatGPT as inputs for emotion classification
, we observe in Table \ref{tab:chatgpt_zsfs} that almost all models that utilize these generated emotion explanations (zero- and few-shot) outperform \textbf{Baseline 1} (i.e., ChatGPT's own direct prediction of dominant emotions). 
Specifically, our best zero-shot model \textbf{CEE-Chat} 
outperformed \textbf{Baseline 1} in exact match accuracy by 20 percent points and top-2 accuracy by 13 percent points.

We also compared the performances with the more complex ChatGPT baseline, (i.e., \textbf{Baseline 2}, where we prompt ChatGPT to generate emotion explanations \textit{and} emotion predictions at once). Our models that utilize ChatGPT-generated explanations consistently outperformed \textbf{Baseline 2} in the zero-shot setting. However, there was no distinguishable improvement in either of the few-shot approaches.

In addition, we observed that models with few-shot generations of emotion explanations perform noticeably better than the ones with zero-shot generations in both exact match and top 2-accuracies. Between the 2 few-shot approaches, the frame ignorant approach yields better emotion classification performance. We suspect that this is due to the class imbalance of the frames in the original dataset.


Additionally, models trained and evaluated on the Concatenated Emotion Explanations in \textit{CEE} show better exact match performances than models trained and evaluated on individual Emotion Explanations in \textit{EE}. However, the performance of \textit{EE} slightly outperformed that of \textit{CEE} in the top-2 accuracies few-shot setting (Table \ref{tab:chatgpt_zsfs}). 

We also conducted McNemar's test \cite{mcnemar1947note}, a statistical hypothesis test used to analyze the significance of differences between paired categorical data. It assesses whether the differences are due to random variation or indicative of a real effect (P-value < 0.05). We performed McNemar's test 
on differences between emotion outputs of models built with only headlines and outputs of models built with ChatGPT-generated emotion explanations. Among the results, model trained with few-shot generated Concatenated Emotion Explanations, without frame information (i.e., the \textbf{CEE-Chat} model) significantly outperforms the prediction of the \textbf{Headline} model (P-value = 0.013). In addition, in terms of the exact match accuracy, the model trained with zero-shot individual Emotion Explanations generation (i.e., the \textbf{EE-Chat} model), shows significant improvement over the \textbf{Headline} model predictions as well (P-value  = 0.024). Both notable boosts prove that GPT-generated explanations are valuable for explaining human's emotion and can act as a temporary replacement for human-generated contents when resources are limited.

Overall, our best performing models: \textbf{CEE-Chat} and \textbf{EE-Chat} with the frame ignorant few-shot ChatGPT generations achieve a top-2 accuracy of 0.85 and an exact match accuracy of 0.66, reaching a similar performance as our upper bound (\textbf{CEE}) while significantly outperforming emotion classification with the headline-only (\textbf{Headline}) model from Table \ref{tab:single_labels}. This demonstrates the potential of leveraging large language models like ChatGPT in low-resource scenarios where there may not be enough headline and human-generated emotion explanation pairs to train models that mimic human-generated emotion explanations.

\begin{table}[htbp]
\tiny
    \begin{center}
    \begin{tabular}{|L{0.12\columnwidth} | L{0.12\columnwidth}| L{0.59\columnwidth}| }
    \hline
      \textbf{Headlines}  & \textbf{Emotions} & \textbf{Generated CEE}  \\ \hline
      Cruz confronted by mother of Santa Fe shooting victim at rally & Truth:\newline sadness \newline\vspace{2mm} \textbf{HEADLINE}:\newline anger  \newline\vspace{2mm} \textbf{CEE-Chat}:\newline sadness & Seeing a mother confront the man who could have prevented her child's death gives me a sense of admiration for her courage and resolve; Why are they confronting at a rally? Is it safe there?; As a parent, I can feel her pain and anguish. Losing a child is unimaginable.; I don't understand what's happening from the headline alone; How terrible it must be for her to face the person she feels is responsible for her child's death; This mother is so strong to face him ; I hope she can find peace after this confrontation; It's about time someone stood up to him, it's unfortunate it had to be a grieving mother; This headline makes me feel sorry for the mother. She must be going through so much right now.; She's so brave, I can't imagine the strength it took for her to confront him.  \\
      \hline
      Trump meets with families of victims of Texas shooting to listen and learn & Truth:\newline contentment \newline\vspace{2mm} \textbf{HEADLINE}:\newline sadness  \newline\vspace{2mm} \textbf{CEE-Chat}:\newline contentment & The president is meeting with the families, it's a start at least; Trump, the man who was backed by the NRA, is now meeting with victims of gun violence. The irony is not lost on me; I hope he genuinely listens to these families and works towards gun control; Trump's meeting with families, but will he actually do anything about it?; I can't help but feel that this is just a publicity stunt for Trump; I hope this experience makes him understand the gravity of gun violence; I am skeptical about this, given Trump's previous stand on gun control. I hope he genuinely learns something from this; It's a positive step but action speaks louder than words. Let's see if there's any real change after this meeting; It's sad that this meeting is even necessary. Gun violence has taken so many lives; I fear that this is just for show and no real change will come from this.  \\
    \hline
    \end{tabular}
    \caption{Examples where our best performing \textbf{CEE-Chat} model (few-shot w/o frames) is able to correctly predict the emotion for headlines where the \textbf{Headline} model made a mistake. }
    \label{tab:cee-chat-vs-headline}
    \end{center}
\end{table}

Table \ref{tab:cee-chat-vs-headline} provides qualitative examples where our best-performing \textbf{CEE-Chat} model correctly predicts the emotions, whereas the \textbf{Headline} model fails. In the first example, the headline insinuates a sense of heightened tension. This potentially misleads the \textbf{Headline} model since it exclusively relies on the headline's textual content. In contrast, the \textbf{CEE-Chat} model's generated emotion explanations offer deeper context, encompassing a broader emotional landscape and thereby ensuring a more holistic understanding. Moreover, by examining the generated emotion explanations, we obtain better insights into why the model predicts a certain emotion. This offers more interpretability of our emotion prediction model, which is an area that many of the neural classification models lack \cite{zafar2021lack, shen2022interpretability}. Among the various ways to leverage this interpretability, one is to check for potential bias or controversies within the model by directly inspecting the generated emotion explanations. The second example in Table \ref{tab:cee-chat-vs-headline} serves as a notable illustration for this. The mention of Trump, a polarizing figure, can introduce biases as reflected in some the generated explanations. These biases could have steered the model to predict negative emotions such as anger. However, the overall dominant emotion (e.g., contentment) in the generated explanations overrides these minor biases and guides the model to predict the correct emotion, demonstrating the model's robustness towards controversial headlines.

\subsection{ChatGPT Emotion Discussion}
\begin{table}[t]
\begin{center}
\begin{tabular}{ | p{0.22\columnwidth}| p{0.22\columnwidth}| p{0.22\columnwidth}|  }
     \hline
     \multicolumn{3}{|c|}{KL Divergence} \\
     \hline
     Zero-shot & Few-shot w/o frames & Few-shot w/ frames \\
     \hline
     1.42 & 0.84 & 0.99 \\
     \hline
\end{tabular}

\caption{Average KL divergence between human annotated emotions and ChatGPT generated (zero-and few-shot) emotions (from \textbf{Baseline 2}).}


\label{tab:alignment}
\end{center}
\end{table}

In addition, to understand the emotion distribution alignment between human and ChatGPT emotions, we computed the average KL divergence across all test headlines between the human-annotated emotion distribution and the ChatGPT generated emotion distribution (from \textbf{Baseline 2}). As shown in Table \ref{tab:alignment}, providing ChatGPT with few-shot examples improves the emotion distribution alignment. Specifically, the frame ignorant few-shot generations are best aligned with our human emotion distribution. This finding aligns with our result in Table \ref{tab:chatgpt_zsfs} which shows that the frame ignorant few-shot generations are best for emotion classification. 

\begin{figure}[h]
\centering
\includegraphics[width=1\columnwidth]{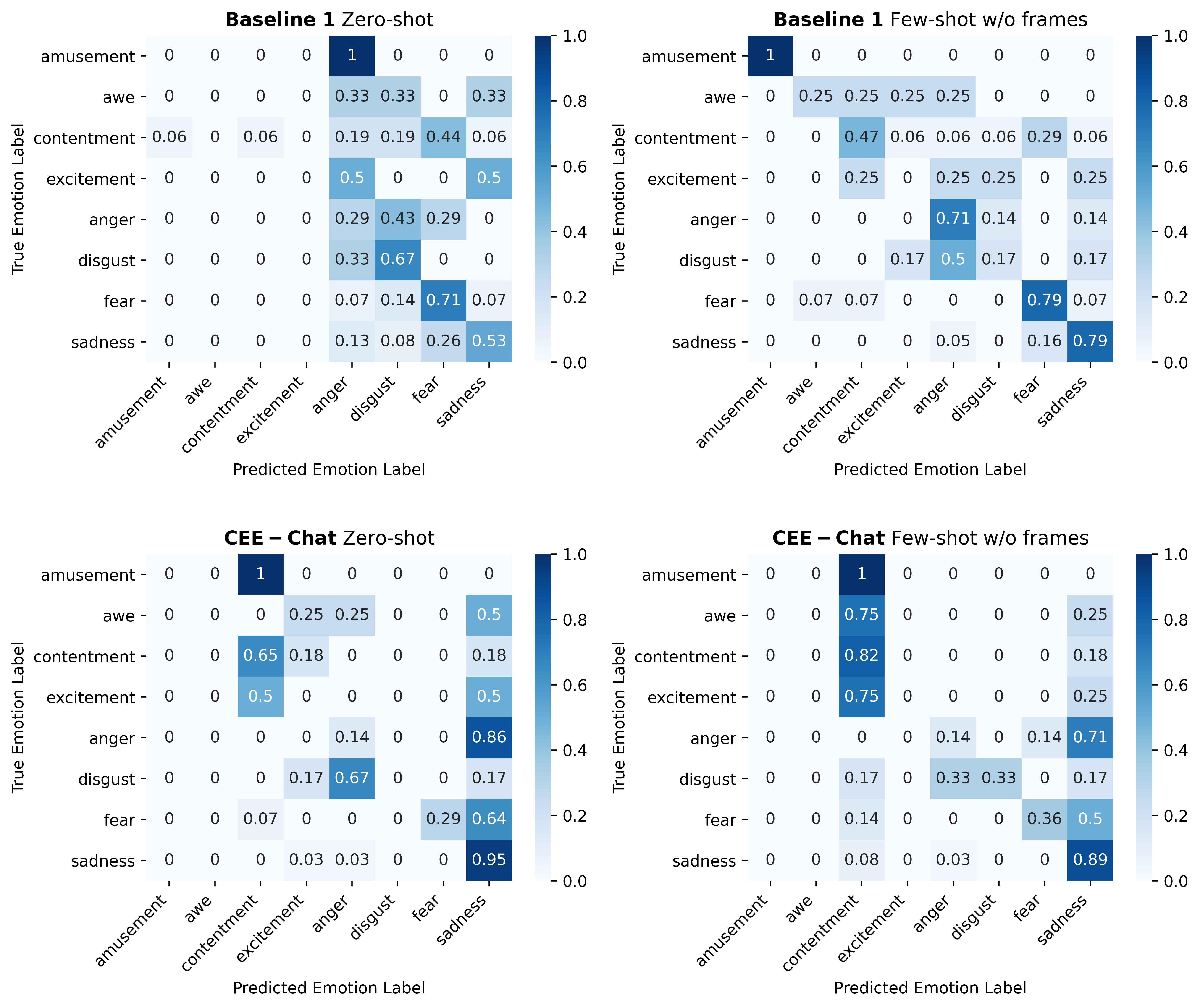}
\caption{Confusion matrices summarizing the prediction results of \textbf{Baseline 1} and \textbf{CEE-Chat}.}
\label{confusion-matrix}
\end{figure}

We also plotted confusion matrices to examine the distribution of predicted emotion labels for each true emotion label (Figure \ref{confusion-matrix}). When comparing the confusion matrices of \textbf{Baseline 1} Zero-shot and \textbf{Baseline 1} Few-shot, we observe that few-shot examples led ChatGPT to generate a wider range of emotions. In the absence of few-shot examples, ChatGPT predominantly generated negative emotions for almost all news headlines related to gun violence. Another notable finding is that ChatGPT struggles to distinguish between anger and disgust. While humans feel different emotions based on the subtle nuances in specific news headlines, ChatGPT falls short in this aspect. We also observe that headlines labeled as Awe was not accurately predicted for any of the news headlines. This discrepancy may arise from different areas of focus between humans and ChatGPT when interpreting the news headlines. For instance, when presented with the headline "Doctors share gun stories, demand action after NRA tells them to ‘stay in their lane’", majority of annotators felt awe towards doctors advocating for the right cause according to our emotion explanations written by human annotators. In contrast, ChatGPT responded with anger, perceiving the NRA's action as disregarding the expertise of doctors.

\section{Conclusion}
In today's interactive and participatory media environment, emotion is a key component that steers engagement and attention towards various digital content. The same is true in journalism where affective news has significantly increased; the technological affordances of social media as well as economic incentives drive news professionals to produce stories that are heightened with emotional cues. Given the accessibility to a wide-range of news online, machine learning techniques can be applied in journalism to augment news professionals' understanding of their audiences' emotional responses to digital news. 

Motivated by the promising results of emotion classification model utilizing human-written explanations of emotions, this study explore diverse approaches to leverage emotion explanations to improve the performance of emotion classification models. Even without directly employing emotion explanations at test time, intermediate task training on emotion explanation dataset can effectively improve the performance of the emotion classification task. Results show that using ChatGPT to generate emotion explanations from headline text helps with emotion classification and outperforms using only the headline text. Additionally, providing ChatGPT with few-shot examples can steer the generations to better align with the human emotions. Further analysis of comparing confusion matrices also shows that prompting few-shot examples to ChatGPT helps alleviate its tendency to predominantly generate negative emotional responses.

For future work, we emphasize the significance of implementing context-enriched datasets, particularly in emotion detection and classification models used in public-interest technologies. The technical methods explored in this study can be applied in practical ways: e.g. in developing a newsroom tool for editors and audience engagement professionals. The tool can be utilized to help mitigate second order consequences of affective news consumption such as hyper-sensationalizing stories, creating echo chambers and filter bubbles and over- or under-representing certain issues or events. 


\section*{Limitations}

This study focused on the task of predicting the dominant emotion when reading news headlines that elicit consistent emotions among general public. Specifically, we assumed the existence of a single ground truth answer corresponding to each headline and evaluated various methodologies in this setting. However, we posit that a singular emotion is not embedded within the headline itself, and acknowledge that emotional responses to headlines may significantly stem from readers' individual backgrounds and beliefs. We ultimately advocate that emotion prediction systems should move towards predicting the distribution of elicited emotions among readers rather than solely predicting the dominant emotion. We aim to explore this direction in future research.

It should also be noted that LLMs including ChatGPT may have biases towards specific news topics. As we are providing ChatGPT with few-shot experiments to diversify the generations for our experiments, concerns about bias in LLMs are lessened. However, if this is not the case, the use of emotion explanations generated by LLMs at test time needs to be reconsidered. Moreover, while we documented the version and the experiment settings, ChatGPT is a proprietary system, which brings challenges for replication.

\section*{Ethics Statement}
Ensuring accurate classification of emotions elicited by news headlines poses significant ethical challenges due to the diverse interpretations and backgrounds of individuals. Previous studies have explored emotion classification using news headlines by collecting emotion data through human participants, but this can be limiting due to the size of participants available and the emotion data following those participants. This study offers a novel way to gather a large set of free-text emotion explanations generated by ChatGPT. While this approach is meant to enhance human capacity and the limitations of data size, we acknowledge that generative data is inherently synthetic and not a transparent representation of human emotional responses. Current large language model systems are not fully transparent in the various types of data used for general-purpose generation, making it difficult to state with certainty, the kinds of biases that may be embedded in the emotional responses.

\section*{Acknowledgements}
This work is supported in part by the U.S. NSF grant 1838193 and DARPA HR001118S0044 (the LwLL program). The U.S. Government is authorized to reproduce and distribute reprints for Governmental purposes. The views and conclusions contained in this publication are those of the authors and should not be interpreted as representing official policies or endorsements of NSF, DARPA, or the U.S. Government.


\nocite{*}
\section{Bibliographical References}\label{sec:reference}

\bibliographystyle{lrec-coling2024-natbib}
\bibliography{
main}


\end{document}